\documentclass[letterpaper, 10 pt, conference]{ieeeconf}  

\IEEEoverridecommandlockouts                              

\usepackage{booktabs}
\usepackage{multirow}
\usepackage[normalem]{ulem}
\useunder{\uline}{\ul}{}
\usepackage{cite}
\usepackage{graphicx} 
\usepackage{siunitx}
\usepackage{subcaption}
\usepackage{xcolor}
\usepackage{tikz}
\usetikzlibrary{shapes.geometric}
\usepackage{comment}
\usepackage{textcomp}
\usepackage{amsmath} 
\usepackage{amssymb}  
\usepackage{todonotes}
\usepackage[breaklinks=true]{hyperref}
\usepackage{gensymb}
\usepackage{algpseudocode}
\usepackage[ruled,vlined]{algorithm2e}
\definecolor{TUMBlue}{rgb}{0.0, 0.40, 0.74}
\definecolor{TUMGray1}{rgb}{0.2, 0.2, 0.2}
\definecolor{TUMGray3}{rgb}{0.8, 0.8, 0.8}
\definecolor{TUMBlue4}{rgb}{0.60, 0.78, 0.92}
\definecolor{TUMIvory}{rgb}{0.85, 0.84, 0.80}
\definecolor{TUMOrange}{rgb}{0.89, 0.45, 0.13}
\definecolor{TUMGreen}{rgb}{0.64, 0.68, 0.0}
\definecolor{TUMGreenWeb}{rgb}{0.62, 0.73, 0.21}
\definecolor{TUMRedWeb}{rgb}{0.92, 0.45, 0.22}
\usepackage{tikz}
\usepackage{textcomp}
\usepackage{lipsum}
\usepackage{fancyhdr} 

\title{\LARGE \bf
OpenLiDARMap: Zero-Drift Point Cloud Mapping using Map Priors}

\author{Dominik Kulmer\textsuperscript{\textdagger}$^{*}$, Maximilian Leitenstern$^{*}$, Marcel Weinmann$^{*}$ and Markus Lienkamp$^{*}$%
\thanks{$^{*}$Dominik Kulmer, Maximilian Leitenstern, Marcel Weinmann, and Markus Lienkamp are with the Institute of Automotive Technology, Munich Institute of Robotics and Machine Intelligence (MIRMI),
Technical University of Munich, 85748 Garching, Germany}%
\thanks{\textsuperscript{\textdagger}Corresponding author: \tt\small dominik.kulmer@tum.de}%
}
\begin{document}

\maketitle
\thispagestyle{empty} 
\pagestyle{empty} 

\pagestyle{fancy} 
\fancyhead{} 
\renewcommand{\headrulewidth}{0pt}  

\fancyhead[L]{\small{Preprint. This work has been submitted for possible publication.}}

\thispagestyle{empty} 
\pagestyle{empty} 

\begin{abstract}
\thispagestyle{fancy}
Accurate localization is a critical component of mobile autonomous systems, especially in Global Navigation Satellite Systems (GNSS)-denied environments where traditional methods fail. In such scenarios, environmental sensing is essential for reliable operation. However, approaches such as LiDAR odometry and Simultaneous Localization and Mapping (SLAM) suffer from drift over long distances, especially in the absence of loop closures. Map-based localization offers a robust alternative, but the challenge lies in creating and georeferencing maps without GNSS support. To address this issue, we propose a method for creating georeferenced maps without GNSS by using publicly available data, such as building footprints and surface models derived from sparse aerial scans. Our approach integrates these data with onboard LiDAR scans to produce dense, accurate, georeferenced 3D point cloud maps. By combining an Iterative Closest Point (ICP) scan-to-scan and scan-to-map matching strategy, we achieve high local consistency without suffering from long-term drift. Thus, we eliminate the reliance on GNSS for the creation of georeferenced maps. The results demonstrate that LiDAR-only mapping can produce accurate georeferenced point cloud maps when augmented with existing map priors. 
\end{abstract}

\section{\uppercase{Introduction}}
\label{sec:introduction}
Localization is essential for mobile autonomous systems, enabling them to navigate and interact with their environment effectively. In recent years, significant advancements have been made in LiDAR odometry~\cite{Vizzo2023,Zheng2024}, LiDAR-inertial odometry~\cite{Bai2022,Xu2022,Zheng2024_2}, and Simultaneous Localization and Mapping (SLAM)~\cite{Dellenbach2022,Yifan2024,Koide2024_2,Pan2024}. Despite their advances, these methods remain prone to drift, especially during long-term operations. While loop closure can help reduce drift by aligning the current position with a previously visited location, it does not guarantee an accurate reconstruction of the intermediate path. Additionally, loops are not always present, especially in linear or open-ended trajectories.

\begin{figure}[]
    \centering
    \includegraphics[width=\linewidth]{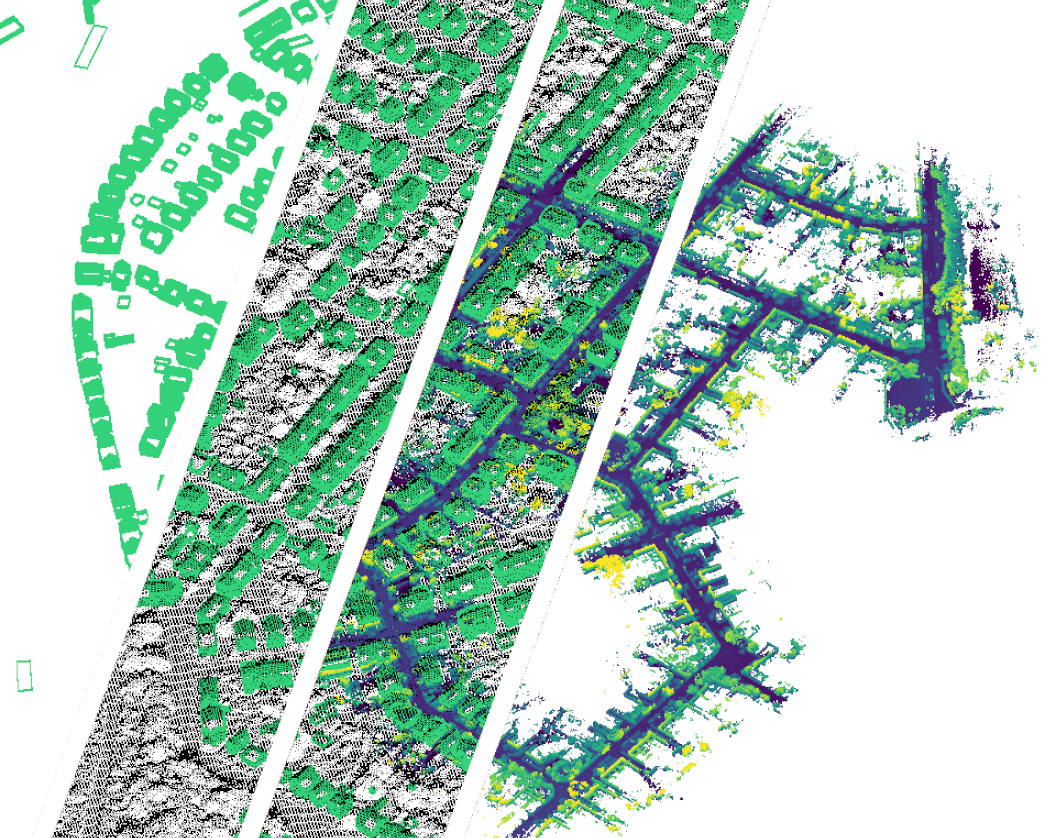}
    \caption{Representation of the different map priors and formats from the building approximations on the left to the final georeferenced point cloud map on the right for KITTI Seq. 00.}
    \label{fig:pipeline}
\end{figure}

Map-based localization offers an alternative by using pre-existing maps to constrain localization and eliminate drift. However, these algorithms rely on the availability of maps, which limits their applicability in unknown environments where LiDAR odometry, LiDAR-inertial odometry, and SLAM excel. OpenStreetMap\footnote{www.openstreetmap.org}~(OSM), a globally available web-based platform, provides maps that can be used for localization with minimal geographic constraints. However, these standard-definition (SD) maps cannot support high-precision localization due to their limited accuracy.

High-definition (HD) maps represent the environment at a resolution of \SIrange[]{10}{20}{\centi \meter}~\cite{Jeong2022} and are ideal for precise localization ~\cite{Koide2019,Koide2024}. These maps typically consist of high accuracy point cloud maps. However, the map generation process suffers from the same drift and accuracy issues as the localization itself.
Global Navigation Satellite Systems (GNSS) offer a global reference to mitigate drift and enable georeferenced mapping. However, their signals are highly susceptible to degradation in obstructed environments. This vulnerability significantly reduces their reliability for producing high-precision maps.

This work addresses these challenges by proposing a method for high-definition map generation that combines publicly available building maps and surface models with onboard LiDAR data. By leveraging these resources, the approach eliminates reliance on GNSS and enables the creation of accurate, georeferenced point cloud maps (\autoref{fig:pipeline}).

The main contribution of this paper is a pose-graph-based optimization algorithm that combines Iterative Closest Point (ICP) scan-to-scan with scan-to-map matching of publicly available building data and sparse surface models to generate accurate georeferenced point cloud maps without the need for GNSS data.
While scan-to-map matching regulates long-term drift, scan-to-scan matching maintains local consistency between individual LiDAR scans. It also enables the approach to compensate for missing buildings and outdated map data or to bridge short areas not present on the sparse prior map.

Our approach uses neither learning-based methods, feature extraction techniques, nor loop closures.
We show that the approach works with a single parameter set on different platforms, like vehicles and segways, with different LiDAR setups, ranging from a single 32-channel LiDAR to a modern multi-LiDAR setup, and in various environments, such as residential and rural regions.

In sum, we make four claims: Our approach is able to (I) map long sequences without accumulating drift over time; (II) automatically georeference the generated map without GNSS data; (III) keep a high local consistency of the generated map; (IV) yield promising results on multiple robotic platforms, LiDAR setups and environments without further tuning.

To build on our work, we provide an open source implementation upon acceptance of the paper.

\section{\uppercase{Related Work}}
\label{sec:related_work}
When available, high-precision RTK-GNSS signals are often used for georeferenced mapping, requiring precise time synchronization between sensors and optimal signal reception along the mapped route. However, GNSS signal interruptions, such as in urban canyons or underpasses, pose a challenge to continuous mapping.
Hybrid pipelines have been proposed to deal with signal interruption. These typically involve creating an initial map using LiDAR odometry or SLAM, followed by post-processing to georeference the map using GNSS data~\cite{Leitenstern2024}. While this approach can bridge short gaps without GNSS, it remains dependent on intermittent signal availability. SLAM techniques fused with GNSS signals can also produce georeferenced maps~\cite{Koide2019,Cramariuc2023,Dellaert2022}, but require careful tuning of sensor weights and suffer from the same dependency on GNSS signal quality.

OSM represents an alternative to GNSS for global localization and mapping, relying on widely available crowdsourced map data.
\cite{Floros2013} improved global localization accuracy by combining visual odometry with Monte Carlo localization, using chamfer matching to align trajectories with OSM maps.
~\cite{Suger2017} proposed a probabilistic navigation method that aligns 3D-LiDAR sensor data with OSM tracks using semantic terrain information and a Markov-Chain Monte Carlo framework.
\cite{Yang2017} introduced a Gaussian-Gaussian cloud model for visual odometry, where OSM road constraints help mitigate drift and resolve scale ambiguities. 
~\cite{Yan2019} used OSM to create orthophoto-style images of roads and building footprints and generated semantic descriptors to match LiDAR data.
~\cite{Ballardini2021} introduced a localization method that detects building facades using stereo image-based point clouds and matches them with 3D building models from OSM.
Similarly, \cite{Cho2022} computed angular distances to buildings within OSM and produced descriptors that match LiDAR data to achieve localization. 
\cite{Elhousni2022} used a particle filter to integrate LiDAR point clouds with OSM constraints such as road boundaries, improving accuracy by exploiting map geometry. 
~\cite{Frosi2023} extended this concept by combining SLAM with OSM priors, integrating 2D building geometry into trajectory estimation through LiDAR scan matching. 

These methods highlight the diverse applications of OSM in reducing drift for mapping and localization in the absence of GNSS signals. However, challenges such as limited accuracy and dependence on map accuracy remain.

Satellite or aerial imagery is another alternative for georeferenced mapping by aligning ground-level observations with overhead map features. \cite{Miller2021} presented a cross-view localization framework that leverages semantic LiDAR point clouds alongside top-down RGB satellite imagery for georeferenced mapping. Similarly, \cite{Xia2024} refined cross-view localization techniques to operate effectively in areas lacking fine-grained ground truth, relying on coarse satellite or map data to improve accuracy. However, the inherent misalignment between ground observations and aerial imagery remains a persistent challenge, often resulting in significant localization errors.

Pure LiDAR odometry and SLAM methods lack georeferencing, while GNSS-based approaches remain vulnerable to environmental interference and signal loss. OSM-based methods, while independent of GNSS, rely on the accuracy of crowd-sourced maps that may be outdated or inaccurate. Satellite data provides broad coverage but often suffers from alignment problems between ground and aerial imagery.

These limitations highlight the need for novel approaches that can achieve georeferenced mapping without relying on GNSS.

\section{\uppercase{Georefenced Point Cloud Mapping}}
\label{sec:georef_mapping}
\begin{figure*}[]
    \centering
    \includegraphics[width=\linewidth]{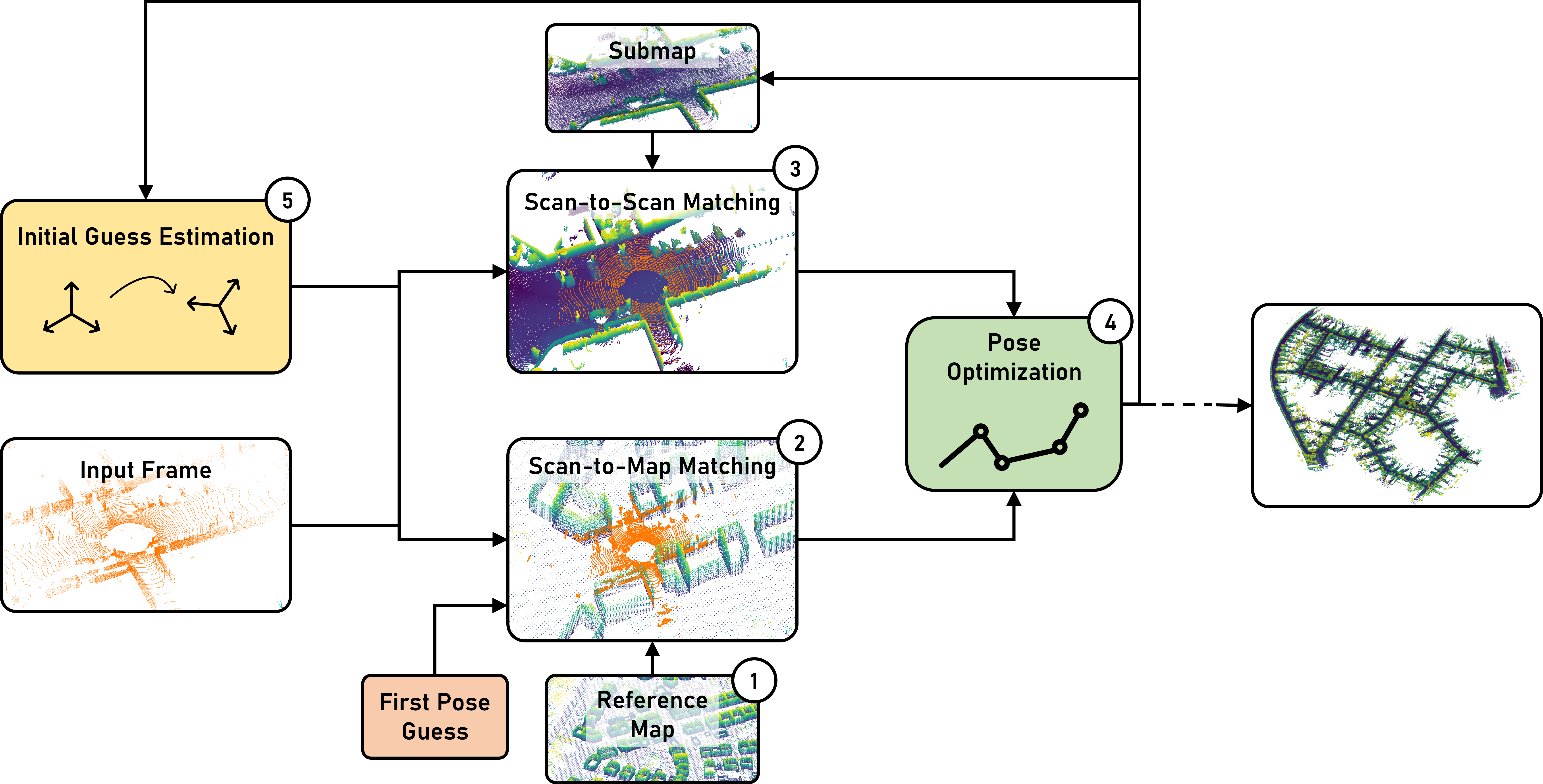}
    \caption{Pipeline overview that shows our main steps to generate georeferenced point cloud maps. Starting with (1) the sparse reference map, which is used for (2) the scan-to-map matching. (3) shows the scan-to-scan matching. The two results are then optimized with (4) a pose-graph optimization, resulting in the final pose, which is then used for (5) the initial guess estimation for the next LiDAR frame.}
    \label{fig:pipeline_diagram}
\end{figure*}

In this section, we explain our LiDAR-based point cloud mapping approach.
The main idea is to combine scan-to-scan matching with scan-to-map matching of reference maps of publicly available building information and sparse surface models in a pose-graph optimization framework. 
The advantage of our method is that we maintain high local consistency while eliminating long-term drift without the need for additional sensors. 
Our mapping procedure can be summarized in five steps (\autoref{fig:pipeline_diagram}); while the first one is done beforehand, the others are performed for each input frame:
\begin{enumerate}
    \item Generating a sparse reference point cloud map from openly available building data and surface models.
    \item Scan matching of the current LiDAR scan and the sparse reference map.
    \item Scan matching of the current LiDAR scan and a local submap of previous LiDAR scans.
    \item Performing a graph optimization of the resulting poses.
    \item Estimating the initial guess for the next LiDAR scan from a constant distance and rotation assumption.
\end{enumerate}

\subsection{Generating Sparse Reference Maps}
\begin{figure}[]
    \centering
    \includegraphics[width=\linewidth]{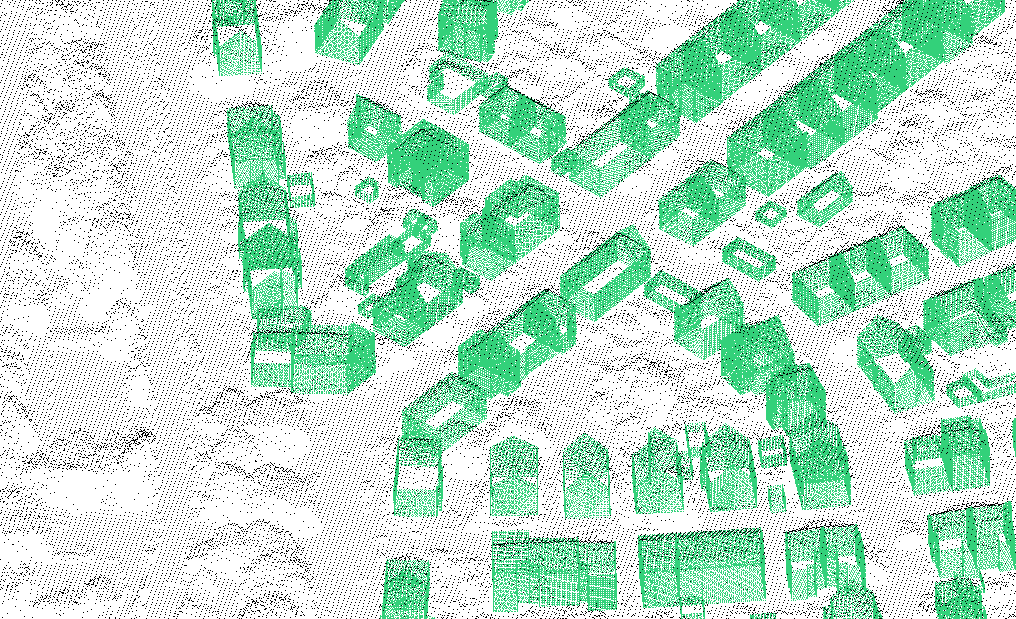}
    \caption{Exemplary illustration of a combined sparse point cloud map of (green) approximated building data and (black) the surface model with a ground sampling distance of \SI{1}{\meter} around the starting position of KITTI Seq. 00.}
    \label{fig:sparse_map}
\end{figure}

OSM provides, besides road infrastructure data, building outlines for much of the world. In addition to geometric outlines, OSM can store semantic attributes such as building height or number of stories. Beyond OSM, other publicly available local data sources, such as Germany's open data portal\footnote{www.govdata.de}, provide detailed building information. 
From this data, sparse georeferenced three-dimensional point clouds can be generated. Depending on data availability, a simplified building representation is derived from OSM data or more sophisticated building models, such as those from the German open data portal. While building models can directly be approximated as a point cloud due to their inherent spatial representation, OSM data requires an estimation for building heights. We assume a height of \SI{4}{\meter} per floor, with a default height of \SI{8}{\meter} for untagged buildings. The resulting point cloud approximates the building shape using a tessellation with an edge length of \SI{0.5}{\meter}.

In addition to building data, three-dimensional surface measurements are available for many regions of the world\footnotemark[2]\footnote{www.data.europa.eu}\footnote{www.data.gov.uk}\footnote{www.usgs.gov}. These surface measurements are typically provided as either surface models or elevation models with varying ground sampling distances, often in the region of \SIrange{1}{5}{\meter}. Surface models capture raw LiDAR scans or a downsampled subset, while elevation models represent the terrain alone, excluding vegetation and structures.
This data is just another representation of point clouds. Therefore, it can be combined with the building point cloud data to create a sparse georeferenced point cloud map (\autoref{fig:sparse_map}). We use this map as a reference in the following scan-to-map matching.

\subsection{Scan-to-Map Matching}
\label{sec:s2m}
Our approach is based on \textit{small\textunderscore gicp}~\cite{Koide2024_3}, a lightweight, header-only C++ library designed for point cloud preprocessing and scan matching. We adopt a voxel hash map as our data structure to efficiently handle large point clouds, allowing fast nearest-neighbor searches. The data within each voxel is stored according to the linear iVox principle~\cite{Bai2022}.

The current input frame is pre-processed with a voxel-based downsampling before the scan matching. The parameters for our entire processing chain can be taken from \autoref{sec:parameters}.

A manual initial pose estimate on the map is provided for the first input frame and subsequently refined through scan-to-map matching. For subsequent frames, the initial estimate is derived from the results of the pose-graph optimization, as described in \autoref{sec:init_guess}.
The scan matching process uses the ICP algorithm for its simplicity and robustness. Outliers are managed using the Geman-McClure robust kernel, and the optimization is carried out using the Gauss-Newton method. The scan matching process returns a global pose within our reference map.

\subsection{Scan-to-Scan Matching}
\label{sec:s2s}
The same procedure used for scan-to-map matching is applied to scan-to-scan matching, using the same pre-processing, data structure, and optimization. 

For scan-to-scan matching, a local submap of the environment is incrementally constructed. The first input frame serves as the registration target for the subsequent scan frame. If the relative transformation between consecutive frames is below \SI{0.1}{\meter}, the system assumes a static state. Therefore, no pose is provided for the optimization, and the optimization process for both scan-to-scan and scan-to-map matching is skipped. In dynamic scenarios, as defined by this threshold, the scan-to-scan matching outputs the relative transformation.

After the pose-graph optimization, the current frame is integrated into the local submap using the determined transformation. To maintain a local representation of the environment, any voxels in the submap that are more than \SI{100}{\meter} from the current position are removed. From the second frame onward, scan matching is performed against this evolving local submap rather than solely against the previous frame. This approach is inspired by KISS-ICP~\cite{Vizzo2023} and leverages the initial pose estimation described in \autoref{sec:init_guess}.

\subsection{Pose-Graph Optimization}
\label{sec:optimization}
Our approach integrates scan-to-scan and scan-to-map matching into an optimization framework using the Ceres solver~\cite{Agarwal2023}. The optimization problem is formulated as a non-linear least squares problem (\autoref{eq:nlls}), where residuals are minimized to align the current pose estimate with the scan matching results. 

\begin{equation}
\min_{x} \frac {1}{2} \sum_{i} \rho _ {i} \left ( ||f_ {i} ( x_ {{i}_ {1}} , \cdots , x_ {{i}_ k}) ||^ {2} \right )
\label{eq:nlls}
\end{equation}

$\rho _ {i}$ represents a loss function to reduce the influence of outliers on the solution.
$f_ {i} (~)$ is the cost function, which depends on the parameter block $( x_ {{i}_ {1}} , \cdots , x_ {{i}_ k})$.

Unlike batch optimization methods, which process all frames simultaneously, the proposed approach employs frame-by-frame optimization. For each frame, the residuals are computed as the deviation between the initial pose estimate and the results obtained from scan-to-scan and scan-to-map matching.
Scan-to-map constraints represent absolute pose estimates derived from aligning the current frame to the sparse map.
Scan-to-scan constraints represent relative transformations between consecutive frames, linking absolute poses through relative measurements. These assemble a classic pose-graph optimization problem.

To mitigate the influence of outliers, residuals are weighted using a robust loss function to reduce the impact of large deviations on the final solution. We use a Tukey loss to aggressively suppress large deviations for the scan-to-map constraints. For the scan-to-scan matching, we use the softer Cauchy loss. These weights reduce the influence of outliers, which are particularly likely in scan-to-map matching due to deviations between the sparse reference map and the current LiDAR frame.
To maintain the reliability of scan-to-map constraints, they are only incorporated into the optimization process if the number of point inliers exceeds \SI{50}{\percent} of the correspondences between the current frame and the sparse map. This criterion ensures that only sufficiently aligned frames contribute to the optimization, mitigating the effect of poor matches.

The optimization framework balances relative constraints from scan-to-scan matching with absolute constraints from scan-to-map matching, providing accurate and drift-corrected pose estimates.
The result of the optimization is a georeferenced pose computed from the results of the scan-to-scan and scan-to-map matching.

\subsection{Estimating the Initial Guess}
\label{sec:init_guess}
We use a constant distance and rotation model for the initial pose estimation. The delta translation between the previous and current position is applied to the current position to estimate the next initial position.
\begin{equation}
\mathbf{t}_{x + 1} = \mathbf{t}_x + (\mathbf{t}_x - \mathbf{t}_{x-1})
\label{eq:trans}
\end{equation}

For the rotation we use a spherical quaternion extrapolation through slerp (\autoref{eq:rot})with a scalar of $s = 2$.
\begin{equation}
{slerp} (\mathbf{q}_{x-1},\mathbf{q}_{x},s)=\mathbf{q}_{x-1}(\mathbf{q}_{x-1}^{-1}\mathbf{q}_{x})^{s}
\label{eq:rot}
\end{equation}

This straightforward method eliminates the need for time stamping across point clouds or individual points, simplifying the implementation. In addition, the lack of a complex kinematic model enhances the versatility of the approach, allowing seamless application across different robotic platforms without the need to modify the methodology.

\subsection{Parameters}
\label{sec:parameters}
We use a single set of parameters for both scan-to-scan and scan-to-map matching, as well as for all subsequent evaluations. This design choice is aimed at simplicity, minimizing the number of parameters and allowing for a more straightforward system configuration.

Each input frame is downsampled with a \SI{1.5}{\meter} voxel filter to reduce the computational load. The voxel maps are constructed with a resolution of \SI{1}{\meter} and can store up to \SI{10}{points} per voxel, maintaining a minimum distance of \SI{0.1}{\meter} between individual points.

For the nearest neighbor search, we consider 27 neighboring voxels within a 3x3x3 cube around the query point. A correspondence threshold of \SI{6}{\meter} is applied to filter out invalid associations during scan matching. For all robust kernels used in the optimization process, we use a static kernel width of \SI{1.0}{} to ensure consistent outlier handling throughout the pipeline.

\section{\uppercase{Experimental Evaluation}}
\label{sec:evaluation}
We present our experiments to show the capabilities of our method. The results of our experiments support our key claims, namely that we can (I) map long sequences without accumulating drift over time; (II) automatically georeference the generated map without GNSS data; (III) keep a high local consistency of the generated map; (IV) yield promising results on multiple robotic platforms, LiDAR setups and environments without further tuning.

\subsection{Experimental Setup}
We evaluate our method on the three different long-distance datasets KITTI Seq. 00~\cite{Geiger2012,Geiger2013}, NCLT 2013-01-10~\cite{Bianco2016}, and EDGAR Campus, each using different robot platforms and sensor setups. The characteristics can be found in \autoref{tab:datasets}.
The KITTI dataset is a widely accepted benchmark for evaluating localization and mapping algorithms. Following the approach demonstrated by IMLS-SLAM~\cite{Deschaud2018} and later adopted by others, such as KISS-ICP~\cite{Vizzo2023} and CT-ICP~\cite{Dellenbach2022}, we apply a correction to the intrinsic calibrations around the vertical axis. Specifically, the point clouds are adjusted by an angle of \SI{0.375}{\degree} to ensure accurate alignment.

The NCLT dataset employs a Segway platform, introducing significant challenges due to the vehicle's high dynamics and diverse environmental conditions. The dataset encompasses outdoor scenes ranging from wide-open parking areas to narrow building passages.

The EDGAR dataset is the longest of the three tested datasets and covers the Technical University of Munich campus in Garching, Germany. It includes a variety of environments, from building complexes and extensive parking areas to country roads, with vehicle speeds ranging from low speeds up to \SI{70}{\kilo\meter \per \hour}. The route traverses both dense and sparsely built-up areas. Unlike KITTI and NCLT, EDGAR features a multi-LiDAR setup, consisting of two spinning and two solid-state LiDARs, which enhances data richness and coverage.
While the first two datasets are freely available, the EDGAR Campus dataset is proprietary. Information about the sensor setup can be found in~\cite{Karle2023,Kulmer2024}.

All vehicles are equipped with RTK-GNSS systems to provide the ground truth data. 

\begin{table}[]
\caption{Caracteristic of the datasets used for the evaluation.}
\resizebox{\columnwidth}{!}{%
\begin{tabular}{@{}lllll@{}}
\toprule \toprule
Dataset                                                  & \begin{tabular}[c]{@{}l@{}}Length\\ {[}m{]}\end{tabular} & \begin{tabular}[c]{@{}l@{}}Frames\\ {[}\#{]}\end{tabular} & \begin{tabular}[c]{@{}l@{}}Sensor \\ Setup\end{tabular}                      & Scenario                            \\ \midrule
\begin{tabular}[c]{@{}l@{}}KITTI\\ \small{Seq. 00}\end{tabular}  & 3724                                                     & 4541                                                      & Velodyne HDL-64                                                              & outdoor, residential, car           \\ \midrule
\begin{tabular}[c]{@{}l@{}}NCLT\\ \small{2013-01-1}\end{tabular} & 1311                                                     & 5120                                                      & Velodyne HDL-32E                                                             & outdoor/indoor, residential, segway \\ \midrule
\begin{tabular}[c]{@{}l@{}}EDGAR\\ \small{Campus}\end{tabular}   & 5289                                                     & 9104                                                      & \begin{tabular}[c]{@{}l@{}}2x Ouster OS1-128\\ 2x Seyond Falcon\end{tabular} & outdoor, rual/residential, car      \\ \bottomrule
\end{tabular}%
}
\label{tab:datasets}
\end{table}

The quality of the maps created cannot be answered with a single metric. A distinction is made between global deviations, such as drift, and local inconsistencies.

\subsection{Global Displacement Evaluation}
We evaluate our approach using the Absolute Trajectory Error (ATE) and the KITTI metric for the Relative Trajectory Error (RTE). The KITTI metric samples the trajectory into segments of varying lengths, ranging from \SIrange{100}{800}{\meter}, and calculates the average relative translational and rotational errors. While the ATE measures the global deviation of the estimated trajectory from the ground truth, the KITTI metric provides insights into the accuracy over medium-length segment pieces.

We compare our approach with three OSM-based global localization and mapping methods and a satellite imagery-based approach for the ATE evaluation. Additionally, we include KISS-ICP as a LiDAR odometry baseline for comparison. However, open-source code is unavailable for all three of the OSM-based methods, and the repository for the imagery-based method is outdated. As a result, we rely on the published data from their respective published works. To enable a comparison with the ground truth, which is positioned in a local coordinate system originating in the first frame, we transform our georeferenced results into the first frame to obtain the same local representation. Then, we calculate the mean and maximum ATE using the ground truth data with \textit{evo}\footnote{https://github.com/MichaelGrupp/evo}.

\begin{figure*}[]
    \centering
    \includegraphics[width=\linewidth]{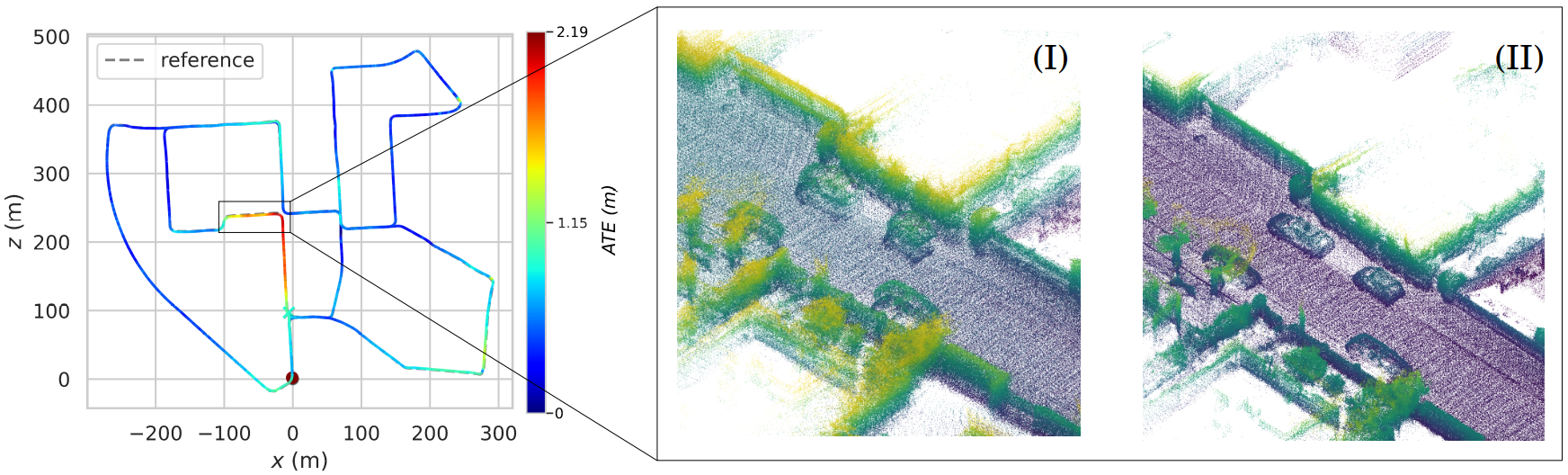}
    \caption{Absolute Trajectory Error (ATE) of our approach for KITTI Seq. 00. (I) Point cloud created from the "ground truth" data. (II) Point cloud created with our approach.}
    \label{fig:kitti_wrong_gt}
\end{figure*}

The results listed in \autoref{tab:ape} demonstrate that our approach reduces the mean ATE by more than half compared to the next best method. Furthermore, our method achieves a mean ATE that is a magnitude lower than the LiDAR odometry approach. \autoref{fig:kitti_wrong_gt} shows that our method accumulates no drift over the traversed distance. Notably, we observe an unusually high deviation in a specific section compared to the rest of the trajectory.
\autoref{fig:kitti_wrong_gt} (I) and (II) show the point cloud generated from the ground truth data and our approach for this section. We identify significant deviations in the provided "ground truth" data for the two passes of the section, particularly with vertical offsets exceeding \SI{1}{\meter} in some regions. 
Hence, the ground truth provided by KITTI cannot be regarded as accurate values, affecting the overall evaluation and the metrics presented.  

\begin{figure}[]
    \centering
    \includegraphics[width=\linewidth]{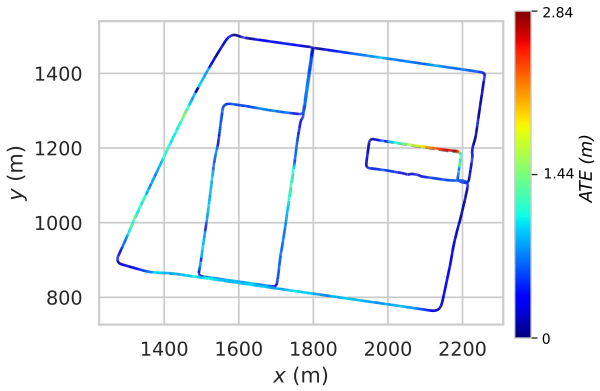}
    \caption{Absolute Trajectory Error (ATE) of our approach for the EDGAR Campus dataset.}
    \label{fig:ape_edgar}
\end{figure}

\begin{figure}[]
    \centering
    \includegraphics[width=\linewidth]{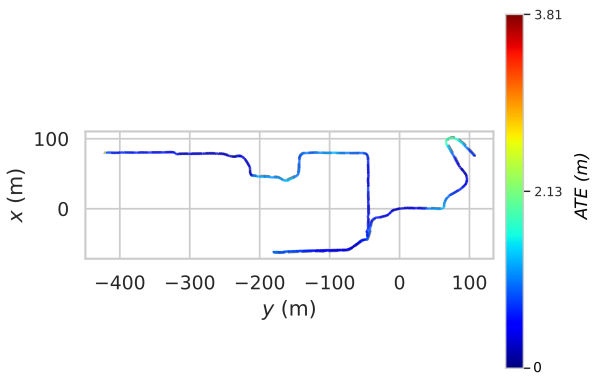}
    \caption{Absolute Trajectory Error (ATE) of our approach for the NCLT 2013-01-1 dataset.}
    \label{fig:ape_nclt}
\end{figure}

Beyond the KITTI dataset, we demonstrate similar performance for the NCLT and EDGAR datasets, where our method exhibits no drift over time. On the EDGAR dataset, again the largest ATEs occur in the section with bad GNSS signals, most likely due to the signal blockage caused by underpasses and tall building walls (\autoref{fig:ape_edgar}). For NCLT, although deviations are the largest, our pose estimation remains robust despite the challenging dynamics and even across an unmapped indoor section (\autoref{fig:ape_nclt}).

\begin{figure}[]
    \centering
    \includegraphics[width=\linewidth]{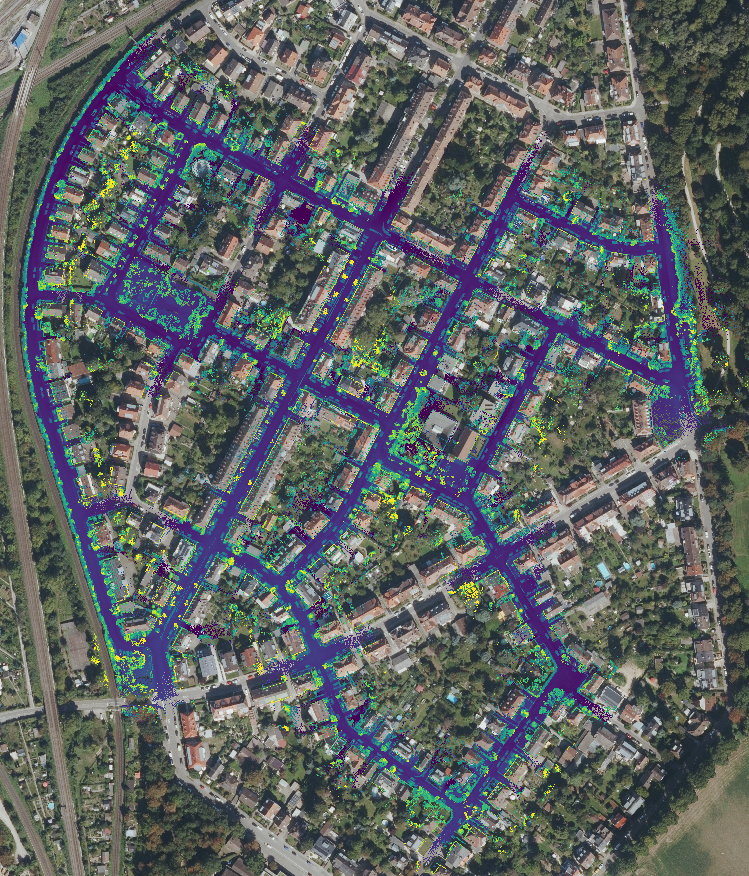}
    \caption{KITTI Seq. 00 point cloud map, created with our approach, plotted to the orthophoto of Karlsruhe.}
    \label{fig:kitti_overload}
\end{figure}

We obtain a global pose directly from our approach and only transform it into a local coordinate system for evaluation. On \autoref{fig:kitti_overload}, we show the overload of the directly generated map by our approach on georeferenced orthophotos of Karlsruhe\footnotemark[2]. 

\begin{table*}[]
\centering
\caption{Mean and max Absolute Trajectory Error (ATE) in [\SI{}{\meter}] for the evaluated datasets. Bold represents the best results and underscores the second-best results.}

\begin{tabular}{@{}ll|llllll@{}}
\toprule \toprule
\multirow{2}{*}{Method} & \multirow{2}{*}{Reference}                & \multicolumn{2}{l}{\begin{tabular}[c]{@{}l@{}}KITTI Seq. 00\end{tabular}} & \multicolumn{2}{l}{\begin{tabular}[c]{@{}l@{}}NCLT 2013-01-1\end{tabular}} & \multicolumn{2}{l}{\begin{tabular}[c]{@{}l@{}}EDGAR Campus\end{tabular}} \\ \cmidrule(l){3-8} 
                        &                                      & mean~\small{$\downarrow$}                                 & max~\small{$\downarrow$}                                  & mean~\small{$\downarrow$}                                 & max~\small{$\downarrow$}                                   & mean~\small{$\downarrow$}                                & max~\small{$\downarrow$}                                  \\ \midrule
RC-MVO                  & \cite{Yang2017}     & 3.76                                 & 14.01                                & -                                    & -                                     & -                                   & -                                    \\
AWYLaI                  & \cite{Miller2021}   & 2.0                                  & 12.0                                 & -                                    & -                                     & -                                   & -                                    \\
LiDAR-OSM               & \cite{Elhousni2022} & {\ul 1.37}                           & {\ul 3.34}                           & -                                    & -                                     & -                                   & -                                    \\ 
OSM-SLAM                & \cite{Frosi2023}    & 3.15                                 & 11.06                                & -                                    & -                                     & -                                   & -                                    \\ \midrule
KISS-ICP                & \cite{Vizzo2023}    & 6.71                                 & 15.20                                & 3.26                                 & 15.99                                 & 66.40                               & 259.13                               \\ \midrule
OpenLiDARMap                  & (ours)                                     & \textbf{0.66}                        & \textbf{2.19}                        & \textbf{1.02}                                 & \textbf{3.81}                                  & \textbf{0.52}                                & \textbf{2.84}                                 \\ \bottomrule
\end{tabular}%

\label{tab:ape}
\end{table*}

In addition to the ATE, we also evaluated the RTE using the KITTI methodology (\autoref{tab:rpe}). For the KITTI dataset, our approach is outperformed by the LiDAR odometry algorithm KISS-ICP. This result can be attributed to the high consistency and smooth trajectory generated by KISS-ICP, which minimizes drift over the short averaged segments used in the KITTI metric. However, on the NCLT dataset, which features high dynamics and a challenging environment, our method demonstrates a better performance. Similarly, the advantages of our approach are evident on the EDGAR dataset, where we maintain consistent trajectories over long distances without drift, leading to lower relative errors.

\begin{table*}[]
\centering
\caption{Relative Trajectory Error (RTE) using the KITTI methodology for the evaluated datasets.}

\begin{tabular}{@{}ll|llllll@{}}
\toprule \toprule
\multirow{2}{*}{Method} & \multirow{2}{*}{Reference} & \multicolumn{2}{l}{\begin{tabular}[c]{@{}l@{}}KITTI Seq. 00\end{tabular}} & \multicolumn{2}{l}{\begin{tabular}[c]{@{}l@{}}NCLT 2013-01-1\end{tabular}} & \multicolumn{2}{l}{\begin{tabular}[c]{@{}l@{}}EDGAR Campus\end{tabular}} \\ \cmidrule(l){3-8} 
 &  & \begin{tabular}[c]{@{}l@{}}trans.~\small{$\downarrow$} \\ {[}\%{]}\end{tabular} & \begin{tabular}[c]{@{}l@{}}rot.~\small{$\downarrow$} \\ {[}deg/m{]}\end{tabular} & \begin{tabular}[c]{@{}l@{}}trans.~\small{$\downarrow$} \\ {[}\%{]}\end{tabular} & \begin{tabular}[c]{@{}l@{}}rot.~\small{$\downarrow$} \\ {[}deg/m{]}\end{tabular} & \begin{tabular}[c]{@{}l@{}}trans.~\small{$\downarrow$} \\ {[}\%{]}\end{tabular} & \begin{tabular}[c]{@{}l@{}}rot.~\small{$\downarrow$} \\ {[}deg/m{]}\end{tabular} \\ \midrule
KISS-ICP & \cite{Vizzo2023} & \textbf{0.51} & \textbf{0.0017} & 2.31 & 0.0161 & 2.77 & 0.0095 \\ \midrule
OpenLiDARMap & (ours) & 0.53 & 0.0025 & \textbf{1.93} & \textbf{0.0116} & \textbf{1.44} & \textbf{0.0046} \\ \bottomrule
\end{tabular}%

\label{tab:rpe}
\end{table*}

To assess how well our approach performs compared to pure scan-to-map algorithms, we tried to evaluate the current state of the art on our sparse reference map. 
However, none of {LiLoc}~\cite{Fang2024}, {BM-Loc}~\cite{Feng2024}, {HDL\textunderscore localization}~\cite{Koide2019}, {DLL}~\cite{Caballero2021}, {FAST\textunderscore LIO\textunderscore LOCALIZATION}\footnote{https://github.com/HViktorTsoi/fast\textunderscore lio\textunderscore localization}, and {lidar\textunderscore localization\textunderscore ros2}\footnote{https://github.com/rsasaki0109/lidar\textunderscore localization\textunderscore ros2} was able to finish KITTI Seq. 00, leaving us without any results to compare.

\subsection{Local Consistency Evaluation}
For quantitative evaluation of map consistency, we use the Mean Map Entropy (MME) metric~\cite{Razlaw2015}. Specifically, we compare the MME between the point cloud map generated by OpenLiDARMap and the map generated from the ground truth transformations of the individual scans.

To facilitate this comparison, we preprocess the point clouds by cropping them to a 100×100~\SI{}{\meter} region and downsampling individual clouds to a voxel size of \SI{0.25}{\meter}. Notably, the composite map remains unfiltered to avoid introducing bias into the evaluation. This preprocessing step is essential to manage the computational complexity caused by the size of the maps.

In the case of the KITTI dataset, we can also use the MME to show that the ground truth is not exact. Our map shows a better (lower) score compared to the ground truth (\autoref{tab:mme}).
The same can also be shown for the NCLT results, where our approach reaches a lower score than the map generated from the ground truth data. For the EDGAR dataset, the MME generated with the ground truth is lower than the results of our approach.

\begin{table}[]
\centering
\caption{Mean Map Entropy for the frames transformed with the ground truth data and OpenLiDARMap results. Lower~\small{$\downarrow$} is better.}
\begin{tabular}{@{}llll@{}}
\toprule \toprule
Method & \begin{tabular}[c]{@{}l@{}}KITTI\\ Seq. 00\end{tabular} & \begin{tabular}[c]{@{}l@{}}NCLT\\ 2013-01-1\end{tabular} & \begin{tabular}[c]{@{}l@{}}EDGAR\\ Campus\end{tabular} \\ \midrule
Ground Truth & -6.293 & -6.212 & \textbf{-6.317} \\ \midrule
OpenLiDARMap (ours) & \textbf{-6.448} & \textbf{-6.339} & -6.231 \\ \bottomrule
\end{tabular}%
\label{tab:mme}
\end{table}

\subsection{Out-of-Date Reference Maps}
Finally, we demonstrate that our method does not rely on up-to-date reference maps. Across all datasets, we show that neither current map data nor temporal alignment between the maps and the LiDAR frames is necessary. For instance, the surface data used in the KITTI dataset originates from 2000 to 2023, while the building data stems from 2023, and the evaluated dataset was collected in 2011. Similarly, the NCLT dataset's data sources span from 2013 to 2024, whereas the EDGAR dataset spans from 2023 to 2024.

\begin{figure}[]
    \centering
    \includegraphics[width=\linewidth]{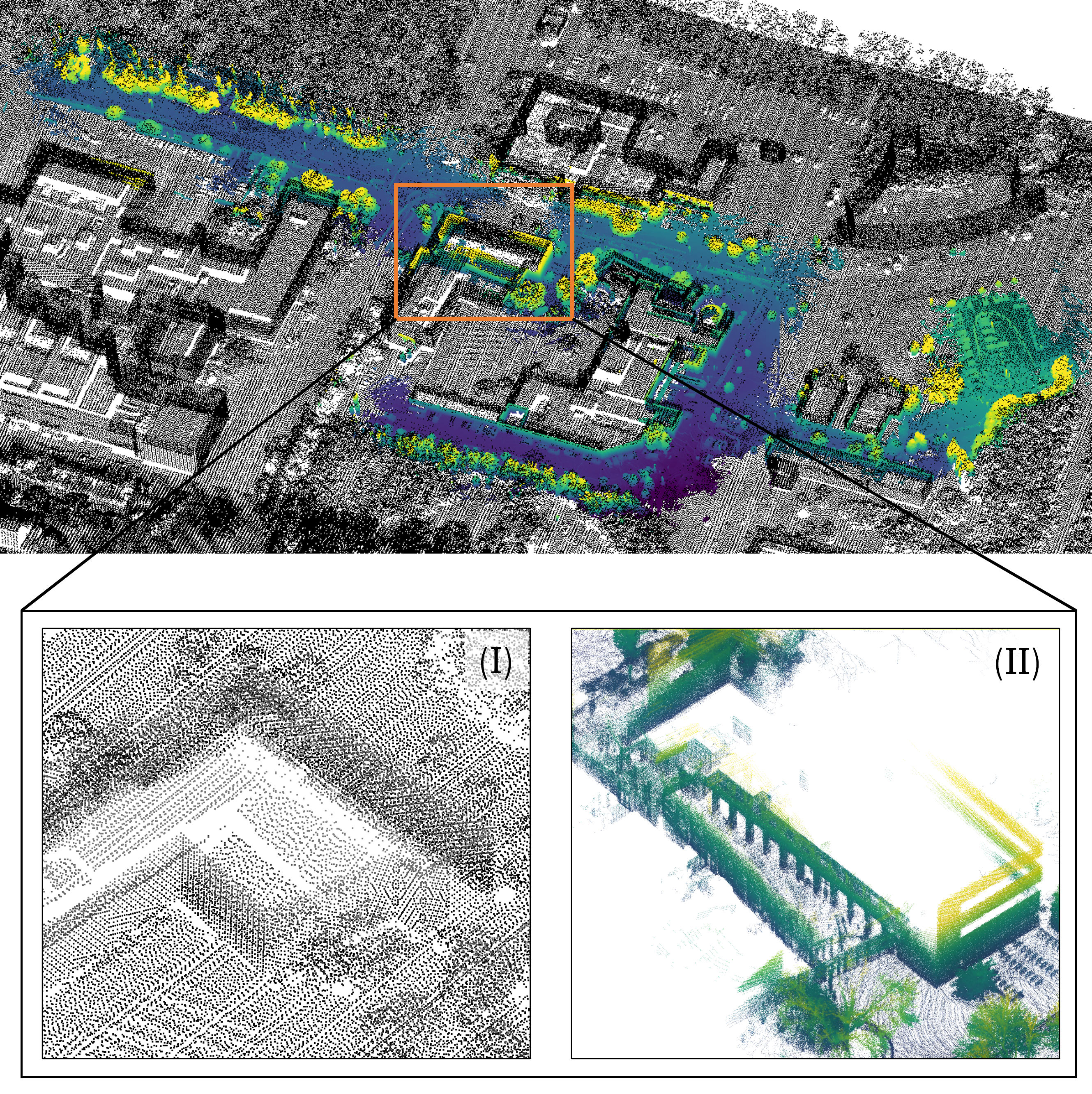}
    \caption{Overload of the sparse reference map and the final point cloud map created with our approach for the NCLT 2013-01-1 dataset. (I) shows part of the reference map, and (II) shows the point cloud map generated with our approach for an indoor section of the dataset that we were able to map despite the missing correspondences between the onboard LiDAR frames and the reference map.}
    \label{fig:nclt_overload}
\end{figure}

\begin{figure}[]
    \centering
    \includegraphics[width=\linewidth]{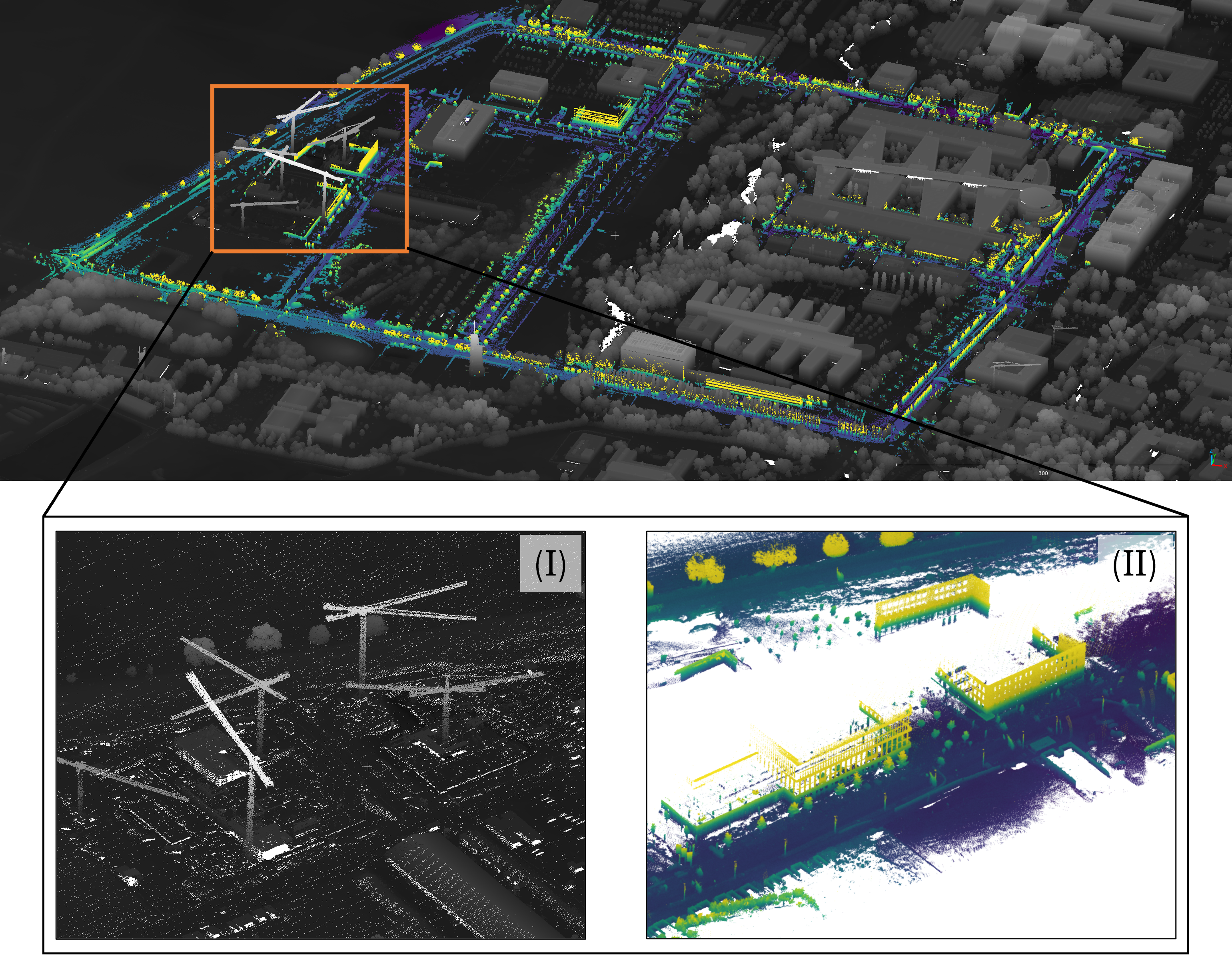}
    \caption{Overload of the sparse reference map and the final point cloud map created with our approach for the EDGAR Campus dataset. Highlighted is a part of the map that still shows (I) the start of construction with multiple cranes in the reference map, while (II) the generated map from the LiDAR frames shows the completed building facades.}
    \label{fig:edgar_overload}
\end{figure}

The accompanying \autoref{fig:nclt_overload} and \autoref{fig:edgar_overload} highlight discrepancies between the sparse maps and the datasets, such as missing buildings or structural differences. These variations emphasize the minimal requirements for input data in our approach. For instance, building outlines for the NCLT dataset were extracted from OSM data, whereas those for KITTI and EDGAR were derived from spatial building models. Moreover, the resolution and type of input data vary significantly: KITTI and NCLT datasets rely on \SI{1}{\meter}-resolution GeoTIFFs, while the EDGAR dataset benefits from raw LiDAR point clouds obtained through aerial scans. This diversity highlights the adaptability of our method to sparse maps with significantly different compositions.

As a final note, our approach focuses on mapping, making runtime not a concern for us. Despite that, we are able to run the entire pipeline in about \SI{30}{\milli\second} per KITTI frame on a modern PC with an AMD Ryzen 7700.

\section{\uppercase{Limitations and Future Work}}
Unlike the ATE analysis, where our approach showed clear improvements, the RTE analysis yielded less conclusive results. This result can be primarily attributed to rotational errors, which have a particularly strong impact on the KITTI metric. Future work could address this limitation by improving the initial estimation, such as using homogeneous transforms or incorporating the time differences between frames.

A key requirement for the functionality of our approach is the availability of a digital surface model. While such data is available for large parts of the world, it is not universally accessible. A notable example is South Korea, where the lack of digital surface models prevents us from evaluating our approach on widely used datasets such as MulRan~\cite{Kim2020} or HeLiPR~\cite{Jung2024}.

Due to the simple ICP-based scan-to-scan matching strategy, our method is limited in handling unmapped regions, such as long tunnels or extensive indoor areas. However, the modular design of our approach allows for future integration of more advanced LiDAR Odometry algorithms to address these challenges. Additionally, significant environmental changes can lead to erroneous results, emphasizing the need for further investigations across diverse scenarios to evaluate robustness under varying conditions.

\section{\uppercase{Conclusions}}
\label{sec:conclusion}
In this paper, we presented an approach for georeferenced point cloud mapping that operates without the need for GNSS. By combining a scan-to-map and scan-to-scan point cloud registration method within an optimization framework, we achieved accurate mapping across diverse environments, vehicles, and LiDAR setups. Our experiments suggest that the proposed method effectively eliminates drift over long distances and is robust to variations in environmental conditions and sensor configurations. Our experiments suggest that the proposed method enables accurate georeferenced point cloud mapping without relying on GNSS sensors.

\section*{\uppercase{Acknowledgements}}
The research was partially funded by the Bavarian Research Foundation (BFS) and through basic research funds of the Institute of Automotive Technology (FTM).

\bibliographystyle{IEEEtran}
\bibliography{export}
\end{document}